\documentclass[letterpaper, 10 pt, conference]{ieeeconf}  % Comment this line out
\IEEEoverridecommandlockouts                              % This command is only
\usepackage{graphics} % for pdf, bitmapped graphics files
\usepackage{epsfig} % for postscript graphics files
\usepackage{times} % assumes new font selection scheme installed
\usepackage{amsmath} % assumes amsmath package installed
\usepackage{amssymb}  % assumes amsmath package installed
\usepackage{color}
\usepackage{url}

\usepackage[utf8]{inputenc}

\newtheorem{prop}{Proposition}

\newcommand{\RR}{{\mathbb R}}

\newcommand{\norm}[1]{\lVert#1\rVert}

\newcommand{\dotex}{{\frac{d}{dt}}}

\title{\LARGE \bf
Accurate 3D maps from depth images and motion sensors
\\ via nonlinear Kalman filtering
}

\author{Thibault Hervier, Silvère Bonnabel, François Goulette
\thanks{Centre de Robotique, Mathématiques et Systèmes, MINES ParisTech, 75272 Paris, France
        {\tt\footnotesize [thibault.hervier]},
       {\tt\footnotesize [silvere.bonnabel]},
       {\tt\footnotesize [francois.goulette]@mines-paristech.fr}}%
}

\begin{document}

\maketitle
\thispagestyle{empty}
\pagestyle{empty}

%%%%%%%%%%%%%%%%%%%%%%%%%%%%%%%%%%%%%%%%%%%%%%%%%%%%%%%%%%%%%%%%%%%%%%%%%%%%%%%%
\begin{abstract}
This paper investigates the use of depth images as localisation sensors for 3D map building. The localisation information is derived from the 3D data thanks to the ICP (Iterative Closest Point) algorithm. The covariance of the ICP, and thus of the localization error, is analysed, and described by a Fisher Information Matrix.  It is advocated this error can be much reduced if  the data is fused with measurements from other motion sensors, or even with prior knowledge on the motion. The data fusion is performed by  a recently introduced specific extended Kalman filter, the so-called Invariant EKF, and is directly based on the estimated covariance of the ICP. The resulting filter is very natural, and is proved to possess strong properties. Experiments with a Kinect sensor and a three-axis gyroscope prove clear improvement in the accuracy of the localization, and thus in the accuracy of the built 3D map.  
\end{abstract}

\section{Introduction}
Accurate 3D mapping from a moving platform and/or localisation in 3D maps has attracted a lot of attention and has become a key issue in the robotics field. Today, the existing methods can be      split into two main groups. Some approaches separate the localisation and mapping processes, either using only the localisation sensors (e.g. in Mobile Mapping Systems \cite{Lara-3D}) or integrating additional perception data in a fusion scheme (e.g. \cite{martinelli}) for the localisation process. Other approaches consider those two processes simultaneously, known as the Simultaneous Localization and Mapping Problem (SLAM methods, e.g. \cite{pathak,dissanayake-2001}). In this paper we address the localisation problem from depth images that arise in mobile robotics, and as a result the construction of accurate 3D maps. 

In order to improve the localisation, and thus the accuracy of the final 3D maps, we propose to combine the information from successive depth images and motion sensors. Although our approach      is quite general, the paper focuses on low-cost sensors,  and presents experiments performed with  a Kinect sensor for the acquisition of depth images, and  three orthogonal gyroscopes as       motion sensors. We advocate  the usual idea at the core of data fusion methods, that assembling complementary sensors and retaining the best part of each, can provide rich information. This idea, however, requires a fine  tuning of the weight of each sensor in the fusion algorithm, and this can only be done if the imperfections of each sensor are well quantified.  

In this paper, we focus on the  ICP (iterative closest point) algorithm as a method for localization from depth images.  The output of the algorithm is a transformation between two clouds of points, and is used as a position sensor. The accuracy of the ICP depends on the sensor's measurement noise, as well as from the richness of the environment. Under some standard assumptions, we compute the covariance of the ICP as in \cite{censi2007accurate,princeton} and we relate it to a Fisher Information Matrix. This computed position and its covariance can then be included in a fusion algorithm. We consider in this paper the point-to-point ICP algorithm, but our method may readily extend to more sophisticated variants, such as point-to-plane ICP. 

The most popular approach for data fusion in mobile robotics is the Kalman filter \cite{Kalman-1961}, which has been proved to be optimal, i.e provide the best estimate possible, for linear systems with      Gaussian white noises. In a non-linear setting, the so-called      extended Kalman filter (EKF) is an extension of the Kalman filter      based on the linearisation of the system. It is only an      approximation of the optimal filter, and its tuning (noise covariances), domain of convergence and stability are still open      issues in the general case \cite{Crassidis}. In this paper, we      propose to use a particular type of EKF, the recently introduced      IEKF \cite{bonnabel_cdc07,bonnabel-martin-salaun-cdc09}, which      accounts for the specific system nonlinearities. This filter is a      particular case of a wider variety of filters specifically      designed for systems on Lie groups,  that have gained increasing      interest over the last years in mobile robotics applications      \cite{bonnabel2009non,hamel-icra07,mahony-et-al-IEEE,Vasconcelos}.      For the considered problem, the proposed filter is new,  is proved      to possess several natural and very remarkable properties,  and is      shown to perform well in experiments.

In Section II it is recalled how the ICP can be used as a pose estimator and how the covariance of the ICP estimate can be computed. In Section III, the results from the ICP are fused with additional data from  motion sensors with an Invariant Extended Kalman Filter. In Section IV  experimental results with a Kinect and a three-axis gyroscope illustrate the benefits of the      approach.

\subsection{Notation}
In order to describe the pose of the robot, we will always
refer to the transformation that maps the mobile frame to
 the ground frame. This transformation can be represented by the homogeneous matrix:
$$
X = \left( \begin{array}{cc}
R & T \\
0 & 1 \\
\end{array} \right)\in\mathrm{SE(3)}
$$
where  $R\in\mathrm{SO(3)}$ is the rotation matrix and  $T\in\RR^3$ the translation vector. We  also define the operator H:  $\RR^3\times\RR^3 \longrightarrow \mathfrak{se}(3)$ (where $\mathfrak{se}(3)$ is the tangent space to SE(3) at identity) that returns the linearized transformation:
$$
H(x_R, x_T) = \left( \begin{array}{cc}
x_R\wedge\cdot & x_T \\
0 & 0 \\
\end{array} \right)
$$

\section{The Iterative Closest Points Algorithm for localisation}

The goal of ICP \cite{besl1992method} in the localisation problem is to estimate the transformation (rotation, translation) that maps one cloud of points to another cloud of points. The  algorithm is iterative in nature, and relies on the following approximation: at each step, it is assumed that each point $p_i$ of the first cloud $Cloud_S$ can be matched to a point $q_i$ of the second cloud $Cloud_T$ such that $q_i$ is the closest point to $p_i$ in $Cloud_T$. Then, a least squares (LS) problem is solved by minimising the cost function
$$
f(X)=\sum_i ||Xp_i-q_i||^2
$$
where X is the rigid transformation that maps $Cloud_S$ to $Cloud_T$.
The solution is constructed from the Singular Value Decomposition of $H=\sum_i p_i q_i^T = U\Sigma V^T$ \cite{caiimplementation}:
$$
R=UV^T \quad, \quad T=C_S-C_TR,
$$
where $C_S$ and $C_T$ are the centroids of the source and target clouds. The essential assumption is that the transformed cloud  is closer to $Cloud_T$ at each new step, making the matching between closest points more relevant. The iterations stop when the convergence is reached.

\subsection{ICP as a pose estimator}

The ICP is often used as a scan matching method only (e.g. \cite{ICP_mapping_ridene}). In this paper, we advocate the  use of this scan-matching information as a pose estimator. Indeed, the algorithm returns an estimation of the transformation between two clouds, hence it provides an estimate of the \textit{relative} displacements of the robot. It can be then used as an absolute pose sensor in several ways:
\begin{itemize}
\item
Either the robot possesses an already built 3D map of the environment, and evaluates at each time its pose with respect to the map via ICP (note that the map can generally be partly constructed from a fixed location before beginning to move).
\item
Or the robot gradually constructs a map of the environment in a fixed ground frame, first by identifying for each cloud its position in the previously constructed map via data fusion, and then by adding the new depth image to the map. This technique is of course subject to drift.
\end{itemize}
\noindent
In both cases, the ICP returns the estimate, denoted Y, of the pose of the robot with respect to the ground frame (i.e the map frame).

\subsection{Sources of error}

There are three main sources of error for the ICP algorithm when matching two clouds \cite{censi2007accurate}, that induce errors on the estimate $Y$:
\begin{enumerate}
\item
Convergence to a bad local minimum : if the initial relative position of the two clouds is not accurate, the matching between closest points is not relevant. Thus the algorithm may converge to a local minimum of the cost function, corresponding to a wrong transformation $X$.
\item
Lack of observability: the clouds might not contain enough spacial information for a perfect localisation. Because of the shapes of the two clouds, a precise matching between them is impossible. For example, it would be impossible to evaluate a movement parallel to a perfectly plane wall with two successive depth images.
\item
Sensor noise: the clouds are inherently noisy so it is impossible to find a perfect match between them.
\end{enumerate}

In this paper, we choose not to consider the case 1) for two reasons. First, because several heuristics, as the ones presented in  Section \ref{experiments:sec}, allow to avoid it in most cases. Then, because such cases pollute the motion estimation. They  thus should be detected and rejected in the data fusion algorithm of Section \ref{kalman:sec}.

\subsection{Fisher information matrix and error covariance}

In order to use the ICP result in a fusion algorithm, it is necessary to be able to quantify the error of the pose estimate given by the ICP, that can be seen as the noisy measurement of the real pose. To do so, we extract the covariance matrix  as follows: 
the problem is  linearised writing
$
X=\delta X X_0
$,
where
\begin{itemize}
\item
$X_0 = \left( \begin{array}{cc}
R_0 & T_0 \\
0 & 1 \\
\end{array} \right)
$ is an approximation of $X$
\item
$\delta X$ is a small rigid transformation in the ground frame.\end{itemize}
Up to second order terms it writes:
$$
\delta X \approx Id+\Omega
\quad \text{with} \quad
\Omega =
\left(
\begin{array}{cc}
\omega\wedge\cdot & \mu \\
0 & 0 \\
\end{array}
\right)
$$

\begin{figure}[b]
\begin{center}
\includegraphics[scale=0.7]{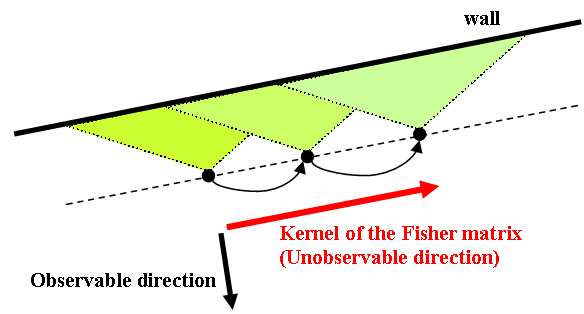}
\end{center}
\caption{Fisher matrix and direction of unobservability}
\label{fisher}
\end{figure}

Thus, the cost function becomes
$f(X) = \sum_i{\norm{ (\delta X X_0) p_i-q_i}^2} = \sum_i{\norm{X_0 p_i-(\delta X^{-1})q_i}^2}
$which implies
$f(X) \approx \sum_i{\norm{X_0 p_i-(I-\Omega)q_i}^2} = \sum_i{\norm{\Omega q_i+(X_0 p_i-q_i)}^2}$.
\\
Let  $A_i$ be the skew matrix associated to a small rotation around  $q_i$:
$$
A_i = (q_i\wedge\cdot) :=
\left(
\begin{array}{ccc}
0 & -q_{iz} & q_{iy} \\
q_{iz} & 0 & -q_{ix} \\
-q_{iy} & q_{ix} & 0 \\
\end{array}
\right)
$$
\noindent
We can now write:
$$
\Omega q_i = \omega\wedge q_i + \mu = -A_i\omega + \mu =
\left(
\begin{array}{cc}
-A_i & I_3 \\
\end{array}
\right)
\left(
\begin{array}{c}
\omega \\
\mu \\
\end{array}
\right)
$$
If we define $ B_i = \left( \begin{array}{cc} A_i & -I_3 \\\end{array} \right)$,
$y_i=X_0 p_i-q_i$ and $x =\left(\begin{array}{cc} \omega & \mu  \end{array}\right)^T$, the cost function finally becomes linear in $x$:
\begin{equation}
f(x) = \sum_i{||y_i-B_ix||^2}
\label{costfunction}
\end{equation}
where $x$ is actually the real state vector that we want to estimate with the ICP. 
Let us consider the two following standard hypotheses \cite{censi2007accurate}: 1) where the two clouds overlap, the closest points assumption induce a true matching between the corresponding sub-clouds.
 2) 
the sensor noise u is a zero-mean Gaussian noise with standard deviation $\sigma$: $u \sim \mathcal{N}(0,\sigma)$. This can be expressed mathematically as 
$\forall_i \ y_i = B_i x+u_i$ with $u_i \sim \mathcal{N}(0,\sigma)$ and
$$
P(y_i|x) = C_i \exp \left(-\frac{\norm{y_i-B_i x}^2}{2\sigma^2} \right)
$$
The noises $u_i$ are assumed to be independent to each other so the log-likelihood of
$P(Y|x) = P(y_1...y_N|x) = \prod_i P(y_i|x)$ is
\begin{equation}
\log[P(Y|x)] = \log(C)-\sum_i \norm{y_i-B_i x}^2/2\sigma^2
\label{probay}
\end{equation}
As the ICP  algorithm  returns the LS estimate $\hat{x}$ of $x$, it is well-known its covariance reaches the Cramer-Rao bound, i.e.,
$$
N = \text{cov}(\hat{x}) = [I(x)]^{-1}
$$
where $I(x)$ is the Fisher information matrix, defined by
$I(x) = -E\left[\frac{\partial^2}{\partial x^2}\log P(Y|x) \right]$.
With (\ref{probay}), it writes
$$
I(x)= \frac{1}{\sigma^2} \sum_i B_i^T B_i
$$
The covariance matrix $N$ of the estimator $\hat{x}$ of $x =\left(\begin{array}{cc} \omega & \mu \\ \end{array}\right)^T$ given by the ICP algorithm is then
\begin{equation}
N = \sigma^2 \left[\sum_i B_i^T B_i \right]^{-1} = \sigma^2 \left[ \sum_i{\left(\begin{array}{cc} -A_i^2 & A_i \\ -A_i & I_3 \\ \end{array}\right)}\right]^{-1}
\label{covR}
\end{equation}
We can observe the direct impact of the two sources of error on the covariance matrix:
\begin{itemize}
\item
The covariance matrix is directly proportional to the variance $\sigma^2$ of the sensor noise.
\item
The Fisher matrix $I(x) \propto \sum_i B_i^T B_i$ truly reflects the spatial information contained in the clouds. If this matrix is singular, its kernel gives the directions of the unobservability (Fig.\ref{fisher}).
Besides, the information matrix is linked to the stability of ICP independently of the noise model, as we have \cite{princeton} :
$
\delta f(x) \propto \delta x^T I(x) \delta x
$
\end{itemize}

If the match between two successive clouds was perfect, we would have had $Y=X \approx \delta X X_0 = (I+\Omega)X_0$. However, we have showed that this observation is not perfect as it involves an estimate $\hat{x}=\left(\begin{array}{cc} \hat{\omega} & \hat{\mu} \\ \end{array}\right)^T$ of $x =\left(\begin{array}{cc} \omega & \mu \\ \end{array}\right)^T$ with cov$(\hat{x})=N$. Thus we can write $\hat{x}=x+v$ where $v =\left(\begin{array}{cc} v_R & v_T \\ \end{array}\right)^T$ is a zero-mean noise of covariance N. 
The noisy observation returned by the ICP is then  
$Y=[I+\hat{\Omega}]X_0 =[I+\Omega+V]X_0$,  with $V = H(v_R,v_T)$  i.e.
\begin{equation}
Y \approx [I+V]X
\label{observationICP}
\end{equation}

\section{Fusion of depth images with motion data via invariant EKF}\label{kalman:sec}

\begin{figure}[t]
\begin{center}
\includegraphics[scale=1]{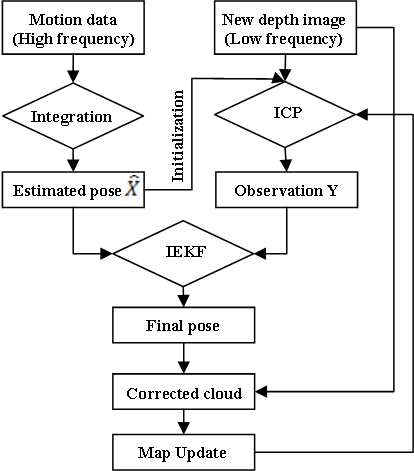}
\end{center}
\caption{Global scheme of the data fusion}
\label{schema}
\end{figure}

In order to improve the ICP estimates, the present paper proposes to fuse them with data from other motion sensors (e.g. odometry, gyroscopes, GPS velocity). The most popular data fusion algorithm  is the extended Kalman filter (EKF) \cite{Crassidis} that has first been used in the Apollo program and has gained popularity in many other fields.  In this paper, we propose to use a specific EKF, the so-called Invariant EKF (IEKF) introduced in \cite{bonnabel_cdc07,bonnabel-martin-salaun-cdc09} that suits particularly well the non-linear structure of the state space. The global scheme of our algorithm is standard, and  is illustrated by  Fig.\ref{schema}.

\subsection{IEKF vs standard EKF}

The kinematic equations of a rigid body moving in space write $\dot{R} = R(\omega\wedge\cdot)$, $\dot{T}=R\mu$ where (R,T) is the pose of the robot, i.e the transformation that maps the body frame to the ground frame, $\omega$ is the instantaneous velocity vector and $\mu$ is the velocity vector, both expressed in the body frame. Using the matrix representation of SE(3) mentioned above, those equation write $\dot{X} = X\Omega$ and the system put in standard state-space form writes:
$$
\begin{cases}
\dot{X} = X\Omega
\\
Y = [I+V]X
\end{cases}
$$
This model is non-linear in nature, as the state-space is represented by matrices with a very particular structure, namely this state-space is a Lie group.

The standard theory of Extended Kalman filter suggests to design an estimate $\hat{X}$ as a dynamical system of the following form
\begin{align}\label{classic:eq}
\dotex{\hat{X}}= \hat{X}\Omega_m + K(Y-\hat{X})
\end{align}
where $\Omega_m$ is the velocity, measured by some sensors, or estimated from prior information on the motion and $K$ is supposed to be tuned via standard KF equations after the system has been linearised. Note that, here $K$ must be a linear function of the entries of its argument, a matrix, i.e. $K$ must be a tensor as well as the covariance matrix, which makes the tuning and the interpretation complicated. Moreover, this construction does not suit the non-linear structure of the state-space : the measurement error $Y-\hat{X}$ involves the error $R-\hat{R}$, which cannot been given a physical meaning because the difference of two rotation matrices is \emph{not} a rotation. As a byproduct,  the estimate equation does not preserve the state-space structure and must be reprojected at each step on the group of SE(3) matrices.

A natural but naive idea to avoid all those problems related to the over parametrization of  SE(3) involved in \eqref{classic:eq} is to work with Euler angles. This is of course possible, but they do not provide a complete and satisfactory parametrization of $SO(3)$, and the space is much distorted around the singularities. Another long standing remedy to this disrespect of space structure that has become standard in the inertial navigation field, is to measure errors  in terms of transformation mapping the estimated rotation $\hat{R}$ to the measurement $R$ i.e errors of the form $\hat{R}R^{-1}$. This leads to multiplicative updates, and it is known as multiplicative EKF (MEKF)  \cite{Crassidis}.

In this paper, we advocate the use of a recent type of EKF, the so-called invariant EKF (IEKF)  \cite{bonnabel-martin-salaun-cdc09}, that extends the idea of the MEKF from $SO(3)$ to arbitrary Lie groups. The general theory was put on firm geometrical ground in \cite{bonnabel2009non} and allows to define sensible observers on Lie groups. Here, the idea of IEKF is simply to define the estimation errors in terms of rigid transformation, and then linearise  the error equation in  a well-chosen frame (the ground frame). Several properties will be detailed in the next subsection.

\subsection{IEKF equations for localization from  depth images}

The IEKF consists of an EKF, but the correction term corrects $\Omega_m$ instead of $\hat{X}$ directly. The correction is applied in the ground frame, yielding the non-intuitive equation :
\begin{equation}
\dotex{\hat{X}} = \hat{X}\Omega_m + E\hat{X},\qquad E =
\left(
\begin{array}{cc}
e_R\wedge\cdot & e_T \\
0 & 0 \\
\end{array}
\right)
\label{dynaestim}
\end{equation}
 where
\begin{itemize}
\item
$\Omega_m$ can be viewed as a noisy measurement of the true velocity $\Omega$ i.e.
\begin{equation}
 \hat{X}\Omega_m = \hat{X}\Omega + W\hat{X}
\label{equaW}
\end{equation}where 
$W = H(w_R,w_T)$ is the drift noise expressed in the estimated ground frame ($w_R$ is the angular velocity noise and $w_T$ on the linear velocity noise).
\item
$E = H(e_R,e_T)$ is a correction matrix based on the discrepancy (i.e the rigid transformation) between the estimated pose and the ICP estimate, as below \eqref{equaE}.
\end{itemize}

Let $\eta = \hat{X}X^{-1}$ be the error defined as the transformation between true and estimated poses, we have
$$
\dot{\eta} = (\dot{\hat{X}})X^{-1} + \hat{X}(\dotex{X^{-1}}) =
 \dot{\hat{X}}X^{-1}-\hat{X}(X^{-1}\dot{X}X^{-1})
$$
Using the kinematic model,  equations (\ref{dynaestim}) and (\ref{equaW})
\begin{equation}
\dotex{\eta} = W \eta + E \eta
\label{equaeta}
\end{equation}
\noindent
By linearising with $\eta = I+\xi+O(\xi^2)$, the error equation in the ground frame becomes
\begin{equation}
\dotex{\xi}= W+E
\label{equaxi}
\end{equation}
where $\xi W$ and $\xi E$ are viewed as second order terms (this assumption can be justified in a stochastic setting \cite{bonnabel-martin-salaun-cdc09}). Letting $\xi = H(\zeta) = H(\zeta_R, \zeta_T)$, $W=H(w)$, (\ref{equaxi}) writes
$$
\dotex{H(\zeta)}=H(w)+H(e)
$$
where $e=H^{-1}(E)$ is the correction vector associated to the matrix E. 
$H$ is a bijection between $\RR^3\times\RR^3$ and $\mathfrak{se}$(3), hence the error vector $\zeta$ follows the equation:
\begin{equation}
\dot{\zeta} = w + e
\label{equazetaint}
\end{equation}

We seek a correction vector e in a form similar to the usual linear Kalman error "$K(Y-\hat{X})$", where K is the Kalman gain (the following token can be rigorously formalized with the Lie Group framework \cite{bonnabel2009non} ):
\begin{itemize}
\item
We want e to vanish when the error $Y\hat{X}^{-1}$ equals the identity matrix (non-linear analogy of the usual case $Y-\hat{X}=0$) thus
$$
(Y-\hat{X}) \longrightarrow (Y\hat{X}^{-1}-Id)
$$
\item
As e is the correction in the $\RR^6$ state-space, the error must be expressed in the same state-space
$$
(Y\hat{X}^{-1}-Id) \longrightarrow H^{-1}(Y\hat{X}^{-1}-Id)
$$
\item
However, $(Y\hat{X}^{-1}-Id) \not\in \mathfrak{se}(3)$ so it is not possible to apply directly the function $H^{-1}$. It is necessary to first apply a projection $\pi$ : $SE(3) \longrightarrow \mathfrak{se}(3)$ such that $\forall M, \ M-Id \approx \pi(M-Id)$, yielding 
$$
H^{-1}(Y\hat{X}^{-1}-Id) \longrightarrow H^{-1}[\pi(Y\hat{X}^{-1}-Id)]
$$
\end{itemize}
Thus the final expression of the correction vector e is
\begin{equation}
e = K*H^{-1}[\pi(Y\hat{X}^{-1}-Id)]
\label{equae}
\end{equation}
The simplest function $\pi$ compatible with the needed properties is:
$$
\pi(M)=\pi(R,T)=(\frac{R-R^T}{2},T)
$$

\noindent
Using $\eta \approx I+\xi$, we have
$Y\hat{X}^{-1} = [(I+V)X]\hat{X}^{-1} = (I+V)\eta^{-1} \approx (I+V)(I-\xi) = I-H(\zeta)+H(v)$. 
Thus 
$e = K*H^{-1}(\pi(Y\hat{X}^{-1}-Id)) \approx K*H^{-1}[Y\hat{X}^{-1}-Id] \approx  K*H^{-1}[-H(\zeta)+H(v)] = K(-\zeta+v)$.
Using (\ref{equazetaint}), we finally obtain the following linearised equation:
$$
\dot{\zeta}=w+Kv-K\zeta
$$
where $v$ is the noise of the ICP, with covariance $N$, and $w$ is the noise of the motion sensor, with covariance, say, $M$. Using the results of linear Kalman filtering theory \cite{Kalman-1961}, the optimal gain K is:
\begin{equation}
K=PN^{-1}
\label{gaincont}
\end{equation}
where P is computed via the continuous Riccati equation
\begin{equation}
\dot{P}=M-P^{-1}NP
\label{Pcont}
\end{equation}
This covariance equation is the same as the one in the case of noisy observations of a constant process, and thus inherits the strong convergence properties of the linear stationary case (i.e., the initial postulated covariance $P(0)$ is exponentially forgot \cite{bougerol}). Finally,  the correction matrix in (\ref{dynaestim}) writes from (\ref{equae}):
\begin{equation}
E = H(e) = H(K*H^{-1}[\pi(Y\hat{X}^{-1}-Id)])
\label{equaE}
\end{equation}

\subsection{Remarkable properties of the filter}

The proposed filter has several sensible properties
\begin{itemize}
\item The filter is based on a measurement error that reflects a true physical discrepancy between the ICP estimate and and $\hat X$. 
\item The estimation $\hat X$ is guaranteed to remain a homogeneous matrix at any time. Moreover, the filter is intrinsically defined, i.e. it does \emph{not} depend on the chosen parametrization of the state. For instance, if rotation matrices are replaced by norm 1 quaternions, the delivered  estimates will be unchanged.
\end{itemize}
The most striking theoretical property of the filter is the following. Consider e.g. Fig.\ref{fisher}. It is clear the ICP will not bring any information along the unobservable direction. As a result it should not affect the estimation $\hat X$ along this direction, i.e. $K$ should be null along this direction. More generally, it means $K$ should tend to align on the Fisher Information Matrix.
\\
However, with a standard EKF, the linearised equation depends on the trajectory $\hat X$ and also on $\Omega$ so there would be no reason why it should tend to the Fisher matrix as $\hat X,\Omega$ vary in time. 
\\
Our observer, on the other hand, plainly benefits from the fact the linearised equation \ref{equazetaint} depends \emph{neither} on $\hat X$ \emph{nor} on $\Omega$, as illustrated by the following proposition, dealing with the convergence properties of the covariance matrix. 
\prop{The Kalman gain admits a very simple interpretation, as for M,N held constant, it converges (even if $\hat X,\Omega$ move) to a stationary gain matrix 
\begin{equation}
K \longrightarrow (MN^{-1})^{1/2}
\label{propK}
\end{equation}
that truly reflects the ratio between the confidence in the model and the confidence in the measurement. In particular if the  noise covariance $M$ is the same in any direction, $K$ tends to be proportional to the Fisher matrix $[I(x)]^{1/2}$, (the more information on a direction, the more corrected it gets).}

\section{Experiments with a Kinect sensor}\label{experiments:sec}

\begin{figure}[t]
\begin{center}
\includegraphics[scale=0.7]{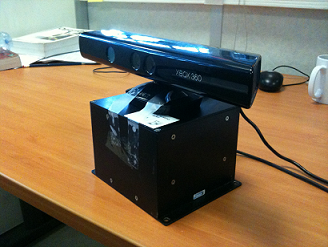}
\end{center}
\caption{Kinect mounted on an IMU}
\label{kinectimu}
\end{figure}

Our goal was to implement an efficient Kalman filter with low-cost sensors. Therefore, we experimented using a Kinect sensor from Microsoft for the depth images and an IMU Crossbow VG600 (Fig.\ref{kinectimu}) for its gyroscopes. The information from the accelerometers, which is generally very noisy and biased, was replaced with prior knowledge on motion uncertainties. Indeed, we opted for the following assumptions on the motion uncertainties :
\begin{itemize}
\item
Because of the approximative spherical symmetry of the gyroscopes, the noise $w_R$ for the rotation vector  is isotropic and its covariance matrix is thus the same in  the mobile or ground frame. The standard deviation on the noise of each gyroscope is around $1 \ deg/s \approx 0.02 \ rad/s$, which yields the diagonal covariance matrix: 
$\text{cov}(w_R) = \text{diag}[(0.02)^2, (0.02)^2,(0.02)^2]$.

\item
During the acquisition, the IMU linear velocity was fluctuating between 0 and $1 \ m/s$ in a horizontal plane and between 0 and $0.5 \ m/s$ vertically (it was carried by a moving person). This yields an approximative uncertainty on the translational motion in the ground frame with $
\text{cov}(w_T) = \text{diag}[(0.5)^2,(0.5)^2,(0.25)^2]
$.
\end{itemize}
\noindent
Finally  the covariance $M$ in the ground frame is:
$$
M = \text{cov}(w) =
\left(
\begin{array}{cc}
\text{cov}(w_R) & 0 \\
0 & \text{cov}(w_T) \\
\end{array}
\right)
$$

\subsection{Implementation}

During our experiment, the IMU had a frequency of 50 Hz whereas the Kinect returned depth images with a frequency of 2 Hz. Therefore, our Kalman filter works as follows:
\begin{itemize}
\item
While a new depth image has not been returned by the Kinect, the gyroscopes are integrated in an open loop to obtain an estimate of the pose. The resulting pose is the prediction $\hat{X}_k^-$ of the filter.
\item
When a new depth image is returned, the corrected pose of the system is computed with the ICP algorithm and the Kalman filter. This yields the updated state $\hat{X}_k^+$.
\end{itemize}
The following steps are repeated every time a new depth image is returned by the Kinect.
\paragraph{High frequency prediction}
Each measurement of the gyroscopes returns the instant rotation vector $\omega_m$, without any information about the translation, thus we let $t_m =0$ and
$$
\Omega_m  = \left(
\begin{array}{cc}
\omega_m\wedge\cdot & 0 \\
0 & 0 \\
\end{array}
\right)
$$
The prediction $\hat{X}_k^-$ of the pose can then be computed by integrating the differential equation $\dot{\hat{X}} = \hat{X}\Omega_m$ for each measure of the gyroscopes, while a new depth image is not returned by the Kinect sensor:
\begin{equation}
\begin{cases}
\hat{T}_{OL} \leftarrow \hat{T}_{OL} \\
\hat{R}_{OL} \leftarrow \hat{R}_{OL}\exp((\omega_m\wedge\cdot) \delta t)
\end{cases}
\label{prediction}
\end{equation}
where $\delta t$ is the time between two consecutive measurements of the IMU ($\delta t \approx 20 \ ms)$. 
These equations are integrated in an open loop from $\hat{X}_{k-1}^+$. When a new image is returned, the prediction state is thus given by $\hat{X}_k^-=\hat{X}_{OL}$.  To decrease the numerical cost,  we used norm 1 quaternions instead of rotation matrices replacing $\hat{R}_{OL} \leftarrow \hat{R}_{OL}\exp((\omega_m\wedge\cdot) \delta t)$ with
$$
\begin{cases}
\hat{q}_{OL} \leftarrow \hat{q}_{OL}+(\frac{1}{2} \hat{q}_{OL}*\omega_m)\delta t \\
\hat{q}_{OL} \leftarrow \text{normalize}(\hat{q}_{OL})\end{cases}
$$
\paragraph{Initialisation of the ICP using depth gradients}
The ICP algorithm measures the variation of pose between the new depth image and the global map constructed from all the previous depth images, yielding the observation Y as described in subsection III-A. To avoid a wrong convergence we propose to improve the initialisation of the relative position of the two clouds. This was done by using features points in order to do a quick matching between the two clouds: after extracting the points with high depth gradient from each cloud, we applied several iterations of ICP between the two resulting sub-clouds. By doing that, we made sure to avoid  wrong convergence in any case, thanks to the sparsity of the two sub-clouds. As the Kinect returns the depth measures row by row, the depth gradients along the horizontal axis can easily be extracted by detecting a jump between two successive depth values.
\paragraph{ICP with the new depth image}
With the previous step, we obtained an estimate of the relative position of the two clouds but the match between the two clouds is far from perfect as we only used a small portion of the points. Thus, it is still necessary to call the ICP algorithm for the registration of the two whole clouds in order to improve the estimate Y. This was done using an already-implemented function of the VTK library \cite{vtk} to perform the ICP (N.B: in order to be able to match clouds that do not entirely overlap, it is necessary to add a patch that deals with too far neighbours).
\\
The resulting covariance matrix $N$ of the noise v of this latter ICP, expressed in the ground frame, is given by (\ref{covR}):
$$
N = \sigma^2 \left[ \sum_i{\left(\begin{array}{cc} -A_i^2 & A_i \\ -A_i & I_3 \\ \end{array}\right)}\right]^{-1}
$$
where $A_i = (q_i\wedge\cdot)$ for each point $q_i$ of the global map and $\sigma \approx 0.2 \ m$ is an approximation of the standard deviation of the noise of the Kinect sensor, for a depth of a few meters.
\paragraph{Computation of the optimal kalman gain K}
In order to calculate the value of the optimal gain K of our Kalman filter, formulas (\ref{gaincont}) and (\ref{Pcont}) must be adapted to the discrete-time case via the standard conversion formulas \cite{salgado1988connection}. The resulting equations are:
$$
K = \frac{P_k(P_k+R)^{-1}}{\Delta t},
\quad
P_{k+1} = Q\Delta t+P_k-P_k(P_k+R)^{-1}P_k
$$
where $\Delta t$ is the time between two successive depth images returned by the Kinect ($\Delta t \approx 500 \ ms$).

\paragraph{Low frequency update}
With the gain K, it is now possible to construct the correction matrix E with (\ref{equaE}):
$$
E = H(K*H^{-1}[\pi([\hat{X}_k^-]^{-1}Y-Id)])
$$
and finally compute the updated pose $\hat{X}_k^+$ after fusion of the data from the ICP and the IMU by integrating the IEKF differential equation $\dot{\hat{X}} = \hat{X}\Omega_m+E\hat{X}$, i.e
\begin{equation}
\begin{cases}
\hat{T}_{k}^+ = \hat{T}_k^- + (e_R \wedge \hat{T}_k^-+e_T)\Delta t\\
\hat{R}_{k}^+ = \exp \left[(e_R\wedge\cdot)\Delta t \right]\hat{R}_k^-
\end{cases}
\label{update}
\end{equation}
The analog formulas using quaternions are
$$
\begin{cases}
\hat{q} = \hat{q}+(\frac{1}{2} \hat{q}*\omega_m) \Delta t +(\frac{1}{2} e_R*\hat{q})\Delta t\\
\hat{q} = \text{normalize}(\hat{q})
\end{cases}
$$

\paragraph{Map update with the new depth image}
From this final pose of the robot, we can compute the real positions of all the points of the depth image in the ground frame, by applying the accurate transformation to the corresponding 3D cloud. The global map is then updated by the addition of these points. 
In our implementation, we managed the global map and the Kinect depth images thanks to the CColouredPointsMap class  of the MRPT library \cite{mrpt}.

\subsection{Results}

\begin{figure*}
\begin{center}
\includegraphics[scale=0.7]{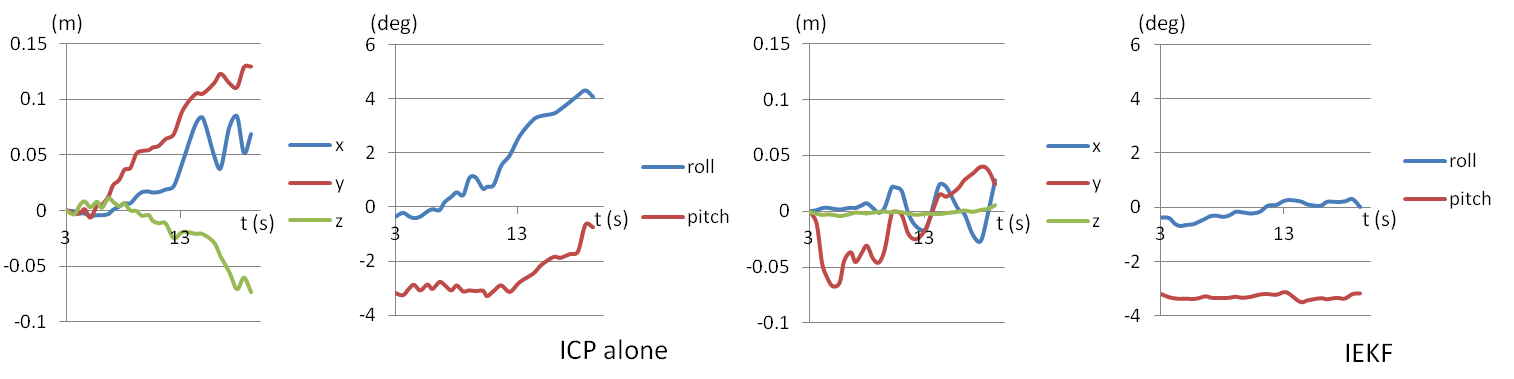}
\end{center}
\caption{Evolution of position/orientation during a rotation around the vertical axis}
\label{graph}
\end{figure*}

\begin{figure*}
\begin{center}
\includegraphics[scale=0.7]{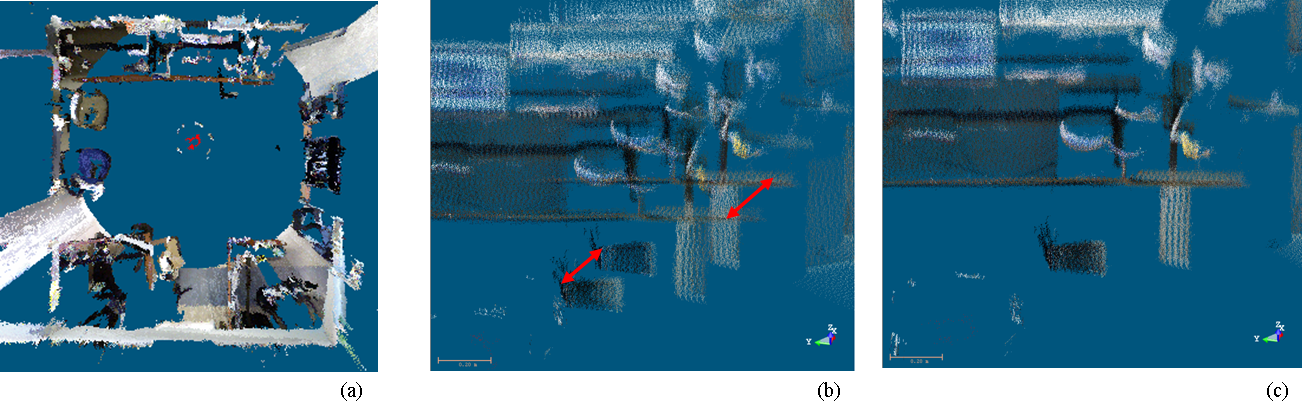}
\end{center}
\caption{360-degrees rotation: (a) bird view, (b) cloud built with ICP only, (c) data fusion with IEKF}
\label{compa}
\end{figure*}

The improvements brought by our method for the localisation problem can be underlined by comparing the results obtained using each sensor on its own with the results from the fusion algorithm, in precise situations.
\\
We first considered a simple situation where the device stayed immobile. Because of the tendency of biases to drift slowly, the integration of the gyroscopes, alone, induced a drift of the attitude. However, thanks to the fusion with the ICP results from two successive depth images, it was corrected by the Kalman filter.
\\
We also compared, for the ICP algorithm alone and for the fusion algorithm, the evolution of the pose of the robot during a (approximate) rotation around the vertical axis (Fig.\ref{graph}). We see that the ICP leads to a drift of the pitch and roll angles (which are supposed to stay constant), which is compensated by a drift of the position variables x,y,z. With the IEKF, this drift has been prevented thanks to the fusion with the gyroscopes. Quantitatively, this correction is underlined by the differences between the computed values of the mean standard deviations of the angle errors : $\sigma_{angle} \approx 5 \ deg$ for the ICP whereas $\sigma_{angle} \approx 1 \ deg$ for the IEKF. We observed the same phenomenon when considering a 360-degrees rotation and comparing the constructed maps (Fig.\ref{compa}). The corresponding acquisition lasted 35s and produced 80 depth images. For the ICP alone, we see an incorrect matching between the initial and the final clouds (image (b)), whereas the final result of the fusion algorithm is a (almost) perfect loop closure (image (c)). 

\section{conclusion}

In this paper, we have demonstrated the potential of depth images as localisation sensors for 3D map building. Using (even low-cost) motion sensors,  the results of ICP are much improved, resulting in an improved accuracy of the built 3D maps. We also advocated that IEKF is a new type of EKF, that suits particularly  well this application. 

In the future, we plan to apply the algorithm described in this paper to other scan matching methods, such as variants of ICP (point to plane ICP) or more complex methods that would also include the color information of the points in the clouds returned by the Kinect.  We also plan to use other 3D sensors, as well as motion sensors such as odometry and/or accelerometers.

%\bibliographystyle{unsrt}
%\bibliography{Article}

\end{document}